\documentclass[runningheads]{llncs}
\usepackage{graphicx}
\usepackage{caption}
\usepackage{subcaption}
\usepackage{enumitem}
\newcommand{\name}{UCNet } 
\newcommand{\vavd}{VAVD }
\newcommand{\fvc}{FVC}

\begin{document}
\title{Misleading Metadata Detection on YouTube}
%
%


\author{Priyank Palod\inst{1} \and
Ayush Patwari\inst{2} \and
Sudhanshu Bahety\inst{3} \and
Saurabh Bagchi\inst{2} \and
Pawan Goyal\inst{1}}

\authorrunning{P. Palod, A. Patwari et al.}

\institute{
\email{\{priyankpalod,pawang.iitk\}@gmail.com}, IIT Kharagpur, WB, India \\ \and
\email{\{patwaria,sbagchi\}@purdue.edu}, Purdue University, IN, USA \\ \and
\email{sudhanshu.bahety@salesforce.com}, Salesforce.com, CA, USA
}

\maketitle              
\vspace{-0.8cm}
\begin{abstract}
YouTube is the leading social media platform for sharing videos. As a result, it is plagued with misleading content that includes staged videos presented as real footages from an incident, videos with misrepresented context and videos where audio/video content is morphed. We tackle the problem of detecting such misleading videos as a supervised classification task. We develop \name - a deep network to detect fake videos and perform our experiments on two datasets - \vavd created by us and publicly available \fvc~\cite{olga_papadopoulou_2018_1147958}. We achieve a macro averaged F-score of 0.82 while training and testing on a 70:30 split of \fvc, while the baseline model scores 0.36. We find that the proposed model generalizes well when trained on one dataset and tested on the other.

\end{abstract}
\section{Introduction}
The growing popularity of YouTube and associated economic opportunities for content providers has triggered the creation and promotion of fake videos and spam campaigns on this platform. There are various dimensions to this act including creating videos for political propaganda as well as choosing clickbaity or shocking title/thumbnails in order to get more views. YouTube itself has classified spam videos into many different categories including misleading metadata (metadata includes the title, description, tags, annotations, and thumbnail).


We use the following definition of ``fake" videos~\cite{papadopoulou2017web}: 
1. Staged videos in which actors perform scripted actions under direction, published as user generated content (UGC).
2. Videos in which the context of the depicted events is misrepresented (e.g., the claimed video location is wrong).
3. Past videos presented as UGC from breaking events.
4. Videos of which the visual or audio content has been altered.
5. Computer-generated Imagery (CGI) posing as real.


Spam detection in social media has been a widely researched topic in the academic community~\cite{wang2010don,viswanath2014towards}. In \cite{koutrika07}, the author describes a model to detect spam in tagging systems. For video sharing platforms, most of the work has concentrated on finding spam comments~\cite{ammari11,radulescu14}.

Recently, there have been works on creating a dataset of fake videos and computationally detecting fake videos \cite{papadopoulou2017web,papadopoulostowards}. As a result, there is a small but publicly available dataset of fake videos on YouTube called the Fake Video Corpus (or FVC) \cite{olga_papadopoulou_2018_1147958}. They had also developed a methodology to classify videos as fake or real, and reported an F-score of 79\% using comment based features. After inspecting their code, we found out that the reported F-score was for the positive (fake) class only. So we reproduced their experiment to find that the macro average F-score by their method is only 36\% since the classifier calls almost all videos as fake. 

Through our experiments, we find that simple features extracted from metadata are not helpful in identifying fake videos. Hence we propose to use a deep neural network on comments for the task and achieve promising results. While using a 70:30 split of the FVC dataset, we find that our method achieves an F-score of 0.82 in comparison to a score of 0.36 by the baseline, and 0.73 by the feature based approach. Further, we also present a new dataset of fake videos containing 123 fake and 423 real videos called VAVD. To see the robustness of our approach, we also train \name on a balanced subset of VAVD, and test on FVC dataset, achieving an F-score of 0.76, better than the score obtained by the feature-based classifier trained on the same dataset. Feature-based classifiers, on the other hand, do not give robust performance while trained on our dataset, and tested on FVC.

\section{Dataset Preparation}



We crawled metadata and comments for more than 100,000 videos uploaded between September 2013 and October 2016 using YouTube REST data API v3. 
The details include video metadata (title, description, likes, dislikes, views etc.), channel details (subscriber count, views count, video count, featured channels etc.) and comments (text, likes, upload date, replies etc.).


With around 100K crawled videos and possibly very low percentage of fake content, manually annotating each and every video and searching for fake videos was infeasible. Also, random sampling from this set is not guaranteed to capture sufficient number of fake videos. Therefore, we used certain heuristics to boost the proportion of fake videos in a small sample to be annotated.

We first removed all the videos with views less than 10,000 (the average number of views in the crawled set) and with comments less than 120 (the average number of comments on a video in the crawled set). This was done to have only popular videos in the annotated dataset.  
A manual analysis of comments on some hand-picked spam videos gave us some comments such as ``complete bullshit'', ``fake fake fake'' etc. Then a search for more videos containing such phrases was performed on the dataset. Repeating the same process (bootstrapping the ``seed phrases'' as well as the set of videos) thrice gave a set of 4,284 potentially spam videos. 
A similar method was adopted for clustering tweets belonging to a particular rumor chain on twitter in \cite{zhao2015enquiring} with good effect. After this, we used the ratio of dislike count:like count of the video for further filtering. Sorting the videos based on the ratio in non-ascending order and taking videos having ratio greater than 0.3 gave us a final set with 650 videos. 

An online annotation task was created where a volunteer was given the link to a video and was asked to mark it as ``spam" or ``legitimate". An option to mark a video as ``not sure'' was also provided. 33 separate surveys having 20 videos per survey (one having only 10) were created and were submitted for annotation to 20 volunteering participants. This task was repeated for a second round of annotation with the same set of annotators without repetition. Statistics from the two rounds of annotation can be seen in Table \ref{tab:anno}. We see that inter-annotator agreement was not perfect, an issue which has been reported repeatedly in prior works for annotation tasks in social media \cite{becker2010learning,ott2011finding}. The discrepancies in the annotations were then resolved by another graduate student volunteer and if any ambiguity still persisted in characterizing the video as spam, the video was marked as ``not sure''. 

We call this dataset as \textbf{VAVD} (Volunteer Annotated Video Dataset) and the annotations\footnote{https://github.com/ucnet01/Annotations\_UCNet} as well as our codes\footnote{https://github.com/ucnet01/UCNet\_Implementation} are publicly available.

\begin{table}[!htb]
\vspace{-0.6cm}
    \caption{Statistics from the two rounds of annotations}
    \begin{subtable}{.5\linewidth}
      \centering
        \caption{Number of videos in different classes}
        \begin{tabular}{|c||c|c||c|}
\hline
\ & Round1 & Round2 & Final \\
\hline
Spam  & 158 & 130 & 123\\
\hline
Legitimate & 400 & 422 & 423\\
\hline
Not Sure & 92 & 98 & 104\\
\hline
\end{tabular}
    \end{subtable}%
    \begin{subtable}{.5\linewidth}
      \centering
        \caption{Annotator agreements}
        \begin{tabular}{|c || c | c | c |}
\hline
 & Spam & Legitimate & Not Sure \\
\hline
Spam & 70 & 62 & 26 \\
\hline
Legitimate & 54 & 308 & 38 \\
\hline
Not Sure & 6 & 27 & 59 \\
\hline
\end{tabular}
    \end{subtable} 
    \label{tab:anno}
    \vspace{-0.3cm}
\end{table}

\textbf{FVC dataset:}
The Fake video corpus (FVC, version 2)\footnote{https://zenodo.org/record/1147958\#.WwBS1nWWbCJ} contains 117 fake and 110 real video YouTube URLs, alongside annotations and descriptions. The dataset also contains comments explaining why a video has been marked as fake/real. Though the dataset contains annotations for 227 videos, many of them have been removed from YouTube. As a result, we could crawl only 98 fake and 72 real videos.
We divide these videos into two disjoint sets, FVC70 (30),  containing 70 (30)\% of these videos for various experiments.


\section{Experiments with simple features}

We first tried using simple classifiers like SVMs, decision trees and random forests on VAVD and test it on FVC, which is the benchmark dataset. We hypothesized several simple features that might differentiate a fake video from a legitimate one, as described below:

\begin{itemize}[leftmargin=*,itemsep=0.05em]

\item has\_clickbait\_phrase: This feature is true if the title has a phrase commonly found in clickbaits. For eg. \textit{`blow your mind', `here is why', `shocking', `exposed', `caught on cam'}. We used 70 such phrases gathered manually.

\item ratio\_violent\_words: A dictionary of several `violent' words like \textit{`kill', `assault', `hack', `chop'} was used. The value of this feature is equal to the fraction of violent words in the title. We hypothesize that violent words generate fear which leads to more views for the video, hence more used in spams.

\item ratio\_caps: This feature is equal to the ratio of number of words in the title which are in upper case to the total number of words in the title.


\item Tweet Classifier Score - Title:  The Image verification corpus (IVC)\footnote{https://github.com/MKLab-ITI/image-verification-corpus/tree/master/mediaeval2016} is a dataset containing tweets with fake/real images. We trained a multi-layer perceptron on IVC to predict the probability of a tweet (i.e., the accompanying image) being fake using only simple linguistic features on the tweet text. Now, we use the same trained network and feed it the title of a video as input. The probability of fakeness that it outputs is then taken as a feature,.

\item dislike\_like\_ratio: Ratio of number of dislikes to number of likes on the video.

\item comments\_fakeness: This is equal to the ratio of comments on the video which mention that the video is fake. To categorize if a comment says that the video is fake or not, we detect presence of words and regexes like \textit{`fa+ke+', `hoax', `photoshopped', `clickbait', `bullshit', `fakest', `bs'}.

\item comments\_inappropriateness: This feature is equal to the ratio of number of comments with swear words to the total number of comments on the video.

\item comments\_conversation\_ratio: This is the ratio of comments with at least one reply to the total number of comments on the video.

\end{itemize}
\vspace{-0.3cm}
Since some of these classifiers are sensitive to correlations in the features, we first decided to remove the lesser important feature among each pair of correlated features. For this, we calculated correlations among the features on all the 100K videos and identified the pair of features with a correlation of more than 0.2. Then we generated feature importance scores using the standard random forests feature selection method and eliminated the lesser important feature from each pair. We trained our classifiers using only these remaining features. Table \ref{tab:results1} shows the performance of some of these classifiers when trained on VAVD and tested on FVC30, as well as when trained on FVC70 and tested on FVC30 with Macro Averaged Precision (P), Recall (R) and F1 Score (F) as the metrics.

\begin{table}[!htb]
\vspace{-1cm}
\begin{small}
    \caption{Performance of simple classifiers tested on FVC30}
     \label{tab:results1}
    \vspace{-0.4cm}
    \begin{subtable}{.5\linewidth}
      \centering
        \caption{training dataset: VAVD}
        \begin{tabular}{@{}c|ccc@{}}
\hline
Classifier          & P & R & F \\ \hline
SVM- RBF            & 0.74          & 0.60       & 0.49        \\
Random Forests		& 0.73			& 0.58		 & 0.46		   \\
Logistic Regression & 0.54          & 0.53       & 0.45        \\
Decision Tree       & 0.53          & 0.52       & 0.46        \\ \hline
\end{tabular}
    \end{subtable}%
    \begin{subtable}{.5\linewidth}
      \centering
        \caption{training dataset: FVC70}
        \begin{tabular}{@{}c|ccc@{}}
\hline
Classifier          & P & R & F \\ \hline
SVM- RBF            & 0.56          & 0.55       & 0.54        \\
Random Forests		& 0.74			& 0.73		 & 0.73		   \\
Logistic Regression & 0.53          & 0.53       & 0.53        \\
Decision Tree       & 0.73          & 0.67       & 0.67        \\ \hline
\end{tabular}
    \end{subtable} 
    \end{small}
   
    \vspace{-0.6cm}
\end{table}



We see that although Random Forests classifier performs the best when trained and tested on FVC, its performance is very bad when trained on VAVD.

To understand the reason for such poor performance of these classifiers, we plotted the PCA of the features on FVC dataset, which is shown in Figure \ref{fig:test} (left). Through the PCA, we can see that though the features may help in identifying some fake videos (the ones on the far right in the plot), for most of the videos, they fail to discriminate between the two classes.

\vspace{-0.4cm}
\section{\name : Deep learning Approach}
\vspace{-0.4cm}

Our analysis during dataset preparation reveals that comments may be strong indicator of fakeness. However, not all comments may be relevant. Hence, we computed ``fakeness vector'' for each comment, a binary vector with each element corresponding to the presence or absence of a fakeness indicator phrase (e.g., ``looks almost real''). We used 30 such fakeness indicating phrases. Now, for each comment, we passed the GoogleNews pre-trained word2vec~\cite{mikolov2013distributed} embeddings of words of the comment sequentially to the LSTM. The 300-dimensional output of the LSTM  is hence referred as ``comment embedding''. We also took the fakeness vector and passed it through a dense layer with sigmoid activation function, to get a scalar between 0 to 1 for the comment called as the ``weight'' of the comment. The idea here was that the network would learn the relative importance of the phrases to finally give the weight of the comment. We then multiplied the 300-dimensional comment embeddings with the scalar weight of the comment to get ``weighted comment embedding''. Now we took the average of all these weighted comment embeddings to get one 300-dimensional vector representing all the comments on the video called the ``unified comments embedding''. The unified comments embedding was then concatenated with simple features described before and passed through 2 dense layers, first with ReLU activation and 4-dimensional output and the second with softmax to get a 2 dimensional output representing the probability of the video being real and fake, respectively. This network is called \textbf{\name} (Unified Comments Net)\footnote{https://bit.ly/2rZ7cAT} and is trained using adam's optimizer with learning rate $10^{-4}$ and cross entropy as the loss function. 

Training the network on VAVD and testing on whole FVC gives an F-score of 0.74 on both classes as shown in Table \ref{tab:tresults2}. Training and testing \name on FVC70 and FVC30 respectively gives a macro F-score of 0.82. We also reproduced the experiments that \cite{papadopoulou2017web,papadopoulostowards} did and found their Macro average F-score to be 0.36 on both 10-fold cross validation and on the 70:30 split.

\begin{table}[!htb]
\vspace{-1cm}
\begin{small}
    \caption{Performance of \name tested on FVC30}
    \label{tab:tresults2}
    \vspace{-0.4cm}
    \begin{subtable}{.5\linewidth}
      \centering
        \caption{training dataset: VAVD}
        \begin{tabular}{c | c c c c}
Class & P & R & F & \#Videos\\
\hline
\hline
Real & 0.64 & 0.88 & 0.74 & 72\\
Fake & 0.88 & 0.64 & 0.74 & 98\\
Macro avg & 0.76 & 0.76 & 0.74 & 170\\\hline
\end{tabular}
    \end{subtable}%
    \begin{subtable}{.5\linewidth}
      \centering
        \caption{training dataset: FVC70}
        \begin{tabular}{c | c c c c}
Class & P & R & F & \#Videos\\
\hline
\hline
Real & 0.74 & 0.87 & 0.8 & 23\\
fake & 0.89 & 0.77 & 0.83 & 31\\
Macro avg & 0.82 & 0.82 & 0.82 & 54\\\hline
\end{tabular}
    \end{subtable} 
    \end{small}
    \vspace{-0.6cm}
    
\end{table}

\begin{figure}[th]
\vspace{-0.5cm}
\centering
\caption{PCA plots. Red dots are Fake Videos while blue dots are Real Videos from FVC.}
\label{fig:test}
\begin{subfigure}{.45\textwidth}
  \centering
  \includegraphics[width=0.8\linewidth]{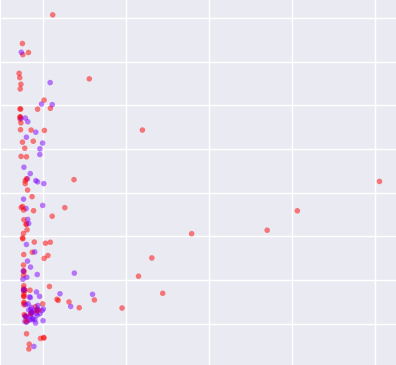}
  \caption{Simple Features}
  \label{fig:PCA_features}
\end{subfigure}%
\begin{subfigure}{.45\textwidth}
  \centering
  \includegraphics[width=0.8\linewidth]{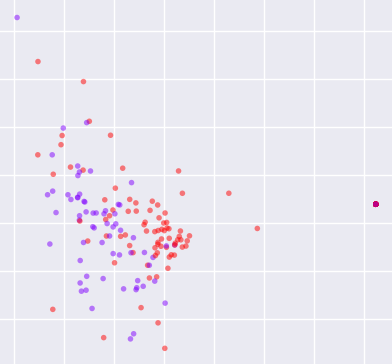}
  \caption{Unified Comment Embeddings}
  \label{fig:pca_unified_comments}
\end{subfigure}
\vspace{-1cm}
\end{figure}

To visualize the discriminating power of comments, we trained \name on VAVD. Then we gave each video of FVC as input to the network and extracted the unified comment embedding. Now we performed PCA of these unified comment embeddings to 2 dimensions and plotted it in Figure \ref{fig:test} (right). 
We observe that comment embeddings can discriminate among the two classes very well as compared to simple features (compare the left and right sub-figures).

Since VAVD had certain properties that overlap with our features (e.g., many fakeness indicating phrases were picked from videos in this dataset), we decided not to test our methods on VAVD as it might not be fair. Hence, we have used it only as a training corpus, and tested on an unseen FVC dataset. Although, future works may use VAVD as a benchmark dataset as well.

Finally, we present Table \ref{tab:comparision} comparing performance of different classifiers when tested on FVC30. Random forests has been reported in the table since it was the best performing simple classifier on the test set. We can see that even if trained on (a balanced subset of) VAVD, \name performs better than any simple classifier or the baseline.

\begin{table}[th]
\vspace{-0.6cm}
\centering
\caption{Overall performance comparison of classifiers on FVC30 test set}
\label{tab:comparision}
\begin{tabular}{c | c | ccc}
\hline
Classifier & Training Set & Precision & Recall & F-Score \\ \hline
\name & FVC70 & 0.82 & 0.82 & 0.82 \\ 
\name & \begin{tabular}[c]{@{}c@{}}class balanced\\ subset of VAVD\end{tabular} & 0.76 & 0.76 & 0.76 \\
Random Forests & FVC70 & 0.74 & 0.73 & 0.73 \\ 
Baseline & FVC70 & 0.29 & 0.5 & 0.37 \\ 
\hline
\end{tabular}
\vspace{-0.8cm}
\end{table}


\section{Conclusions}

Our work presents VAVD, a new dataset for research on fake videos, and also presents \name, a deep learning based approach to identify fake videos with high accuracy using user comments. Future work will involve putting more emphasis on content and metadata than the comments, to be able to detect latest or `breaking news' spam videos.

\section{Acknowledgement}

This material is based in part upon work supported by a Google Faculty Award to Saurabh. Any opinions, findings, and conclusions or recommendations expressed in this material are those of the authors and do not necessarily reflect the views of the sponsor.

 \bibliographystyle{splncs04}
\bibliography{sample-bibliography}

\begin{thebibliography}{10}
\providecommand{\url}[1]{\texttt{#1}}
\providecommand{\urlprefix}{URL }
\providecommand{\doi}[1]{https://doi.org/#1}

\bibitem{ammari11}
Ammari, A., Dimitrova, V., Despotakis, D.: Semantically enriched machine
  learning approach to filter youtube comments for socially augmented user
  models. UMAP pp. 71--85 (2011)

\bibitem{becker2010learning}
Becker, H., Naaman, M., Gravano, L.: Learning similarity metrics for event
  identification in social media. In: Proceedings of the third ACM
  international conference on Web search and data mining. pp. 291--300. ACM
  (2010)

\bibitem{koutrika07}
Koutrika, G., Effendi, F.A., Gy{\"o}ngyi, Z., Heymann, P., Garcia-Molina, H.:
  Combating spam in tagging systems. In: Proceedings of the 3rd international
  workshop on Adversarial information retrieval on the web. pp. 57--64. ACM
  (2007)

\bibitem{mikolov2013distributed}
Mikolov, T., Sutskever, I., Chen, K., Corrado, G.S., Dean, J.: Distributed
  representations of words and phrases and their compositionality. In: Advances
  in neural information processing systems. pp. 3111--3119 (2013)

\bibitem{ott2011finding}
Ott, M., Choi, Y., Cardie, C., Hancock, J.T.: Finding deceptive opinion spam by
  any stretch of the imagination. In: Proceedings of the 49th Annual Meeting of
  the Association for Computational Linguistics: Human Language
  Technologies-Volume 1. pp. 309--319 (2011)

\bibitem{papadopoulostowards}
Papadopoulos, S.A.: Towards automatic detection of misinformation in social
  media

\bibitem{papadopoulou2017web}
Papadopoulou, O., Zampoglou, M., Papadopoulos, S., Kompatsiaris, Y.: Web video
  verification using contextual cues. In: Proceedings of the 2nd International
  Workshop on Multimedia Forensics and Security. pp. 6--10. ACM (2017)

\bibitem{olga_papadopoulou_2018_1147958}
Papadopoulou, O., Zampoglou, M., Papadopoulos, S., Kompatsiaris, Y., Teyssou,
  D.: Invid fake video corpus v2.0 (Jan 2018). \doi{10.5281/zenodo.1147958},
  \url{https://doi.org/10.5281/zenodo.1147958}

\bibitem{radulescu14}
Radulescu, C., Dinsoreanu, M., Potolea, R.: Identification of spam comments
  using natural language processing techniques. In: Intelligent Computer
  Communication and Processing (ICCP), 2014 IEEE International Conference on.
  pp. 29--35. IEEE (2014)

\bibitem{viswanath2014towards}
Viswanath, B., Bashir, M.A., Crovella, M., Guha, S., Gummadi, K.P.,
  Krishnamurthy, B., Mislove, A.: Towards detecting anomalous user behavior in
  online social networks. In: 23rd USENIX Security Symposium (USENIX Security
  14). pp. 223--238 (2014)

\bibitem{wang2010don}
Wang, A.H.: Don't follow me: Spam detection in twitter. In: Security and
  Cryptography (SECRYPT), Proceedings of the 2010 International Conference on.
  pp. 1--10. IEEE (2010)

\bibitem{zhao2015enquiring}
Zhao, Z., Resnick, P., Mei, Q.: Enquiring minds: Early detection of rumors in
  social media from enquiry posts. In: Proceedings of the 24th Int. Conference
  on World Wide Web. Int. WWW Conferences Steering Committee (2015)

\end{thebibliography}


\begin{thebibliography}{8}
\bibitem{ref_article1}
Author, F.: Article title. Journal \textbf{2}(5), 99--110 (2016)

\bibitem{ref_lncs1}
Author, F., Author, S.: Title of a proceedings paper. In: Editor,
F., Editor, S. (eds.) CONFERENCE 2016, LNCS, vol. 9999, pp. 1--13.
Springer, Heidelberg (2016). \doi{10.10007/1234567890}

\bibitem{ref_book1}
Author, F., Author, S., Author, T.: Book title. 2nd edn. Publisher,
Location (1999)

\bibitem{ref_proc1}
Author, A.-B.: Contribution title. In: 9th International Proceedings
on Proceedings, pp. 1--2. Publisher, Location (2010)

\bibitem{ref_url1}
LNCS Homepage, \url{http://www.springer.com/lncs}. Last accessed 4
Oct 2017
\end{thebibliography}
%
\end{document}